\documentclass[letterpaper, 10 pt, conference]{ieeeconf}  

\IEEEoverridecommandlockouts                              

\usepackage{amsmath, amssymb}
\usepackage{graphicx}
\begin{document}
\newpage
\textsuperscript{\textcopyright}2020 IEEE.Personal use of this material is permitted.Permission from IEEE must be obtained for all other uses, in any current or future media, including reprinting/republishing this material for advertising or promotional purposes, creating new collective works, for resale or redistribution to servers or lists, or reuse of any copyrighted component of this work in other works.
\newpage

\title{\LARGE \bf
 A New Approach for Tactical Decision Making in Lane Changing: Sample Efficient Deep Q Learning with a Safety Feedback Reward  
}

\author{Ugur Yavas, Tufan Kumbasar and Nazım Kemal Ure
\thanks{Ugur Yavas is with Eatron Technologies, Istanbul, Turkey {\tt\small ugur.yavas@eatron.com}}%
\thanks{Tufan Kumbasar is with Control and Automation Engineering Department, Istanbul Technical University, Turkey {\tt\small kumbasart@itu.edu.tr}}%
\thanks{Nazım Kemal Ure is with  Artificial Intelligence and Data Science Research Center and Department of Aeronautical Engineering, Istanbul Technical University, Turkey
    {\tt\small kure@itu.edu.tr}}%
\thanks{This work was supported by the Research Fund of the Scientific and Technological Research Council of Turkey under Project 118E807.}
}


\maketitle
\thispagestyle{empty}
\pagestyle{empty}

\begin{abstract}
 
Automated lane change is one of the most challenging task to be solved of highly automated vehicles due to its safety-critical, uncertain and multi-agent nature. This paper presents the novel deployment of the state of art Q learning method, namely Rainbow DQN, that uses a new safety driven rewarding scheme to tackle the issues in an dynamic and uncertain simulation environment. We present various comparative results to show that our novel  approach of having reward feedback from the safety layer dramatically increases both the agent's performance and sample efficiency. Furthermore, through the novel deployment of Rainbow DQN, it is shown that more intuition about the agent's actions is extracted by examining the distributions of generated Q values of the agents. The proposed algorithm shows superior performance to the baseline algorithm in the challenging scenarios with only 200000 training steps (i.e. equivalent to 55 hours driving). 

\end{abstract}

\section{INTRODUCTION}

There has been a growing interest in self-driving cars by the industry since Darpa Urban Challenge \cite{boss}. Despite the great achievements in this competition, the deployment of self-driving cars into production is a quite complicated problem due to reasons such as long tail of edge cases, safety verification and the need of intelligent algorithms that are capable of negotiating with human drivers. There are already level-2 capable cars in production that autonomously control the vehicle at both the longitudinal and lateral levels. However, there is still a need for advancements to level-2 systems, namely the inclusion of automated lane change functionality which is crucial as it covers most of the aspects of highway driving. Thus, we believe that making tactical decisions to change lanes requires intelligence in the context of understanding the behavior of other traffic participants and strict safety monitoring considering the fact that a large amount of accidents happened during this maneuver \cite{lanechangeacc}.

\subsection{Related Work}

The automated lane change problem has been widely handled and various approaches such as rule-based \cite{rulebaedlane}, data-driven supervised learning \cite{borelli}, utility-based \cite{volvorule} and reinforcement learning-based \cite{ilkrl},  to solve the automated lane change problem. However, excluding reinforcement learning, the main drawback of these approaches is the fact that there is no involvement in learning. Thus, these approaches are prone to errors due to noise and uncertainty when the environment changes slightly from the intended design. On the other hand, data-driven algorithms have problems when facing cases outside of their training distribution. Recently, applications of Deep Reinforcement Learning (DRL) to the lane change problem have been investigated by using Q-masking to integrate high-level knowledge \cite{mukadam}, combining with the safety layer \cite{bmw}, injecting uncertainty \cite{itsc}, introducing spatial in-variance with Convolutional Neural Network (CNNs) \cite{volvo1} and combined planning \cite{volvo2}. DRL based methods have clear advantages over other methods considering the fact that they can handle well with uncertainty, measurement noise and large input spaces \cite{drlsurvey}.

\begin{figure}[htb]
    \centering
    \includegraphics[width=1\columnwidth]{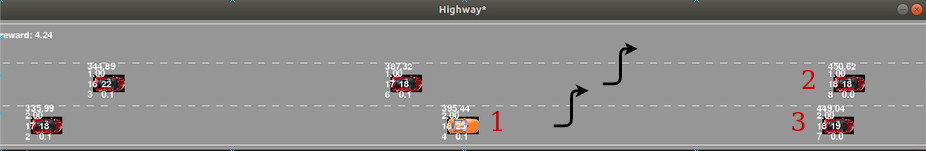}
    \caption{Three lane scenario with orange car (1) being the ego-vehicle: Changing one lane to the left does not bring any speed gain since the lead vehicle in the centre lane (2) is slightly slower than vehicle 3. This scenarios is a common pitfall for the rule based and narrow sighted systems.   }
    \label{fig:training agents}
\end{figure}

The efficient design and implementation of DRL agents involves many steps which are starting with state-action representations, balancing multi-objective reward function, tuning the hyper-parameters of the optimization algorithm, deciding the network architecture, generating rich data out of realistic scenarios and finally broad evaluation against a proper baseline methods with different seeds. Considering the aforementioned steps, \cite{mukadam} lacks the comparison with a fair baseline and uses a very naive simulation environment without challenging scenarios. On the other hand, \cite{bmw} proposes compact state representations that would work in any lane-vehicle number configuration and integrates a safety layer-based on time to collision evaluations of the leader and follower vehicles. The defined safety layer can reject the actions proposed by Q network if it is evaluated as unsafe. Although compact state representation accelerates training (i.e. reduces the amount of computations), it has been underlined in \cite{volvo2} that deciding lane changes by just considering adjacent lanes fails to solve the case shown in Fig. \ref{fig:training agents}. Furthermore, it is stated in \cite{volvo1} that designing a DRL agent that is capable of jointly decide longitudinal and lateral actions performs better than the agent with only making lane change decisions. In \cite{volvo2}, a realistic simulation environment which contains measurement noise, randomized agent behaviors, was used to train a Monte Carlo Tree Search (MCTS) based agent without the consideration of safety.

\subsection{Contribution}

This paper proposes a method to train a sample efficient Rainbow DQN \cite{rainbow} agent that not only makes tactical decisions in the dynamic, uncertain and noisy highway scenarios but also considers safety constraints. The highlights of our contributions can be summarized as:

\begin{itemize}
\item{Implementation of Rainbow DQN to the lane change problem, that results with a major performance increase over double DQN \cite{doubleQ}. It creates more intuition about agents actions by analyzing Q value distributions (See section V). }

\item{The novel use of safety layer that provides a reward feedback to the agent which dramatically increases both sample efficiency and final performance as well as simple but yet efficient safety layer implementation that is aligned with current in vehicle technology such as blind spot warning. Our approach differs from \cite{bmw} as we prefer to use a different safety metric and feed the rejection information from the safety layer as a negative reward in the learning of the agent.}

\end{itemize}

Our results demonstrate that the design and deployment of the Rainbow DQN agent with safety feedback performs significantly better than both rule based and double DQN agents in complex scenarios with different seeds(a,b,c) involving 20 surrounding vehicles having uncertain behavior and reaches to this superior performance only after 200000 training steps.

The paper is organized as follows: Section II gives the details of how DRL methods have been applied to the automated lane change problem. Section III provides the details of the simulation environment and scenario configuration. Section IV shares the results of training and evaluation runs, section V discusses the results and section VI derives conclusion and proposes future research directions.

\section{PROBLEM STATEMENT}

Automated lane change can be formulated as a DRL problem with continuous input state and discrete actions. Actions of the agent need to be evaluated by the safety layer then passed to the low-level controllers that eventually determines the desired steering angle and acceleration. The  ego-vehicle is assumed to have an accurate perception system that could give the relative velocity and position of the other agents. In order to make the perception assumption more realistic, we consider cases with compact state representation and measurement noise. Ideally, perception systems suffer from occlusions, but this was not considered in this study. We also avoid to add longitudinal control actions to the DRL agent since a realistic Adaptive Cruise Control (ACC) system has a lot more complicated design than Intelligent Driver Model (IDM) \cite{idm} and giving DRL agent an extra degree of freedom in the action space would require additional safety verification which is again outside the scope of this work.

\subsection{Reinforcement Learning}

Reinforcement learning is a machine learning paradigm that relies on self-learning agents driven by a reward function which is calculated through interactions with the environment. In every time step, feedback from the environment is received as $S_{t}$ and an action $A_t$ is being selected by the agent and another feedback is received as $S_{t+1}$ with the reward $R_{t+1}$ and future discount $\gamma_{t+1}$. The aforementioned units form a tuple $\langle S, A, T, R, \gamma \rangle$ that is being used to model the Markov Decision Process (MDP) \cite{babuska}. In the model-free reinforcement learning setup, transition function, $T(s, a, s') = P[S_{t+1}=s'\mid S_t=s,A_t=a]$ , is not known and agent tries to find best action set (policy: $\pi$) that maximizes reward without knowing the dynamics of the environment. The problem formulation for the finite horizon $H$ is described as follows:

\begin{equation}
\label{eq:target}
\pi^{*}=\arg \max _{\pi} \mathrm{E}\left[\sum_{t=0}^{H} R_{t}\left(S_{t}, A_{t}, S_{t+1}\right) | \pi\right]\end{equation}
\subsection{Rainbow DQN}

Q-learning is a value-based technique to solve the problem in  (1) by recursively estimating the optimal action-value function $Q^*(s,a)$ \cite{dqn}. By calculating the Q value of each possible state, the action pair and using the Bellman equation \cite{qlearning}, the optimal policy can be attained with a greedy policy of choosing actions with maximum Q values. 

\begin{equation}
\label{eq:qlearn}
    Q^*(s,a) = \mathbb{E}\left[r + \gamma \max_{a'} Q^*(s',a')|s,a\right]
\end{equation}

However,  conventional Q-learning algorithm cannot handle environments having large, continuous state-action pairs. Starting with DQN algorithm \cite{dqn} that approximates Q function with neural networks and uses large experience replay buffers to break correlations in the training data, DRL agents reach outstanding performance in many different tasks \cite{silver}, \cite{levine}.

The state of the art algorithm is the Rainbow DQN \cite{rainbow} which has combined the most significant enhancements over the initial DQN algorithm in terms of training dynamics, sample efficiency and performance. We briefly summarize the main elements of Rainbow DQN below and encourage interested readers to check out the original paper. 

\begin{itemize}
\item 
Double Q learning decouples value estimation and action selection between target.
\item Prioritized experience replay samples more frequently the experience that has bigger loss to speed up the training.
\item Duelling network has the neural network architecture of shared encoder, followed by separate fully connected layers to predict advantage and value of the states separately.
\item Multi-step learning unrolls the equation \ref{eq:qlearn} N-step further to make $Q(s,a)$ values converge faster. 
\item Noisy network adds a noise parameter with normal distribution to the each weight in the fully connected layer which are updated via back-propagation. This leads better exploration strategy than standard $\epsilon$-greedy method. 
\item The $Q(s,a)$ values are predicted as distributions by minimizing the Kullback-Leibner loss. Distributional Q function gives more insight while evaluating a particular state as also shown in figures \ref{fig:turnleft}, \ref{fig:tactical}. 
\end{itemize}

\subsection{State/Action Representations}

In this paper, the agent has three available discrete actions that are keeping the current lane, changing lanes to the left and right which are generated. Regardless of the selected action, IDM handles the longitudinal control and determines following distance and speed. We have used the ego-centered relative state representation which includes the positions and velocities of the other vehicles. However, we have also analyzed the influence of more compact representation as suggested in \cite{bmw} by only providing the information of lead and following vehicles in each lane.

\begin{table}[h]
    \caption{Ego-centring, normalized Cartesian state representation} 
    \vspace*{-0.5mm}
    \centering    
    \begin{tabular}{l c}       
        \hline  \hline \vspace*{-2mm}&\\
        $s_1$, & Normalized ego vehicle speed $v_{ego}/v_{ego}^{d}$ \\
        $s_2$, & Normalized ego vehicle lateral position  $y_{ego}/y_{max}$ \\
        $s_{2i+1}$, & Normalized relative position of vehicle $i$, $\Delta s_i/\Delta s_{max}$ \\
        $s_{2i+2}$, & Normalized relative velocity of vehicle $i$, $\Delta v_i/v_{max}$ \\
        $s_{2i+3}$, &  Normalized relative position of vehicle $i$, $\Delta y_i/y_{max}$ \\
        \hline                                              
    \end{tabular}
    \label{table:Observation_vector_ego}                                
\end{table}

\subsection{Reward Function with Safety Feedback}

In  literature, simple reward functions are defined and used, such as (1) punishing lightly each lane change to limit the number of attempts (2) punishing heavily the accidents and rewarding agent proportional to the target speed. We argue that such a simple reward scheme with the length of the short episode of 1000m may not reflect what the agent has actually learned. In this context, we propose the following novel rewarding scheme combined with game-like episode definition:

\[
    r(s,a,s^\prime)=\left\{
                \begin{array}{ll}
                  \text{speed incentive:   }(v_{current} - v_{initial})/v_{d}\\
                  \text{lane change penalty:  } -1\\
                  \text{if collision then:   } -100(terminal)\\
                  \text{if $v_{current}=v_{d}$ then:   } +100(terminal)\\
                  \text{if action is unsafe then: } -1 \\
                \end{array}
              \right.
  \]
Combining the above reward scheme with longer episode length (5km in our case) would intuitively evaluate the training and evaluation performance of the agent. Instead of using infinite episodes without termination, we consider episodes that are well aligned with the actual use case of automated lane change functionality. A new episode begins whenever the speed of the vehicle gets lower than the desired speed, and thus the agent is expected to make tactical decisions in order to reach the desired set-point (i.e. target speed) once again. During an episode, the intermediate speed of the ego vehicle does not matter if it is settled at a slower speed than the desired ego vehicle speed.

As a second improvement to the rewarding scheme, we propose a novel reward feedback from the safety layer.  In the classical safety approach as in \cite{bmw}, decisions of DRL agents are rejected by a safety layer, and the next action proposed by the agent is evaluated by safety again until an acceptable action is obtained. We enhance this approach in a way to make the safety layer interact with the DRL agent over the reward function. Thus, every time the DRL agent violates safety, it gets -1 reward and the corresponding action is overwritten by the safety layer and the agent receives the information regarding the next state. Proposed safety layer is simple as rejecting the actions that would result in clear accidents such as trying to change lanes while the adjacent lanes are occupied. This technique avoids frequent terminal states by accidents especially during the early stages of training and significantly increases the training speed and the agent's performance.

\subsection{Network Architecture and Training Parameters}

In this paper, we propose two architectures based on Double DQN and Rainbow DQN and compare them with a rule-based agent driven by the Minimizing Overall Braking Induced by Lane Changes (MOBIL) \cite{mobil} algorithm. Both proposed algorithms have networks with CNN layers in the standard implementation to process images from the game environment. Although we are working with continuous measurements not pixels, in order to get a significant performance boost, we have used the CNN layers as proposed in \cite{volvo1}. Following the CNN layer, considering the findings from \cite{bottleneckQ}, large fully connected layers with 256 neurons are being used to prevent over-fitting in the training phase of the networks.

\begin{table}[htb]
\caption{Rainbow Hyper-parameters} 
\vspace*{-2mm}
\centering    
\begin{tabular}{l c}   
\hline  \hline \vspace*{-2mm}&\\
priority replay beta:    & $0.6$  \\
beta schedule steps:    & $100000$  \\
N-step prediction: & $2$ \\
replay size: & $50000$ \\
target network sync freq.: & $500$ \\
learning rate: & $0.0001$ \\
discount factor,$\gamma$: & $0.99$ \\
batch size: & $32$ \\
\hline  
\end{tabular}

\label{table:Highway}                                
\end{table}

\section{SIMULATION ENVIRONMENT}

In this paper, we use the simulator which is the enhanced version of the one presented in our previous work \cite{itsc}. All vehicles except the ego-vehicle are driven by a combination of IDM and MOBIL algorithm. The vehicle motions are defined with the kinematic bicycle model. There are low level longitudinal and lateral controllers that calculate the required acceleration and steering angle of the vehicles. There are four major improvements over the simulator given in \cite{itsc}:
\begin{itemize}

\item{Always block the lane of ego-vehicle with slower vehicle}
\item{During the episodes if every lane is locked by slow vehicles, randomly speed up the slower vehicles}
\item{Randomly select driver profiles from a uniform distribution according to driver table III}
\item{Inject realistic position and velocity measurement noise \cite{radarnoise} to the ego-vehicle states.}
\end{itemize}

IDM \cite{idm} is the standard car-following model that calculates the required acceleration response to reach desired velocity set-point or following distance when there is a lead car . The dynamics of IDM are as follows:

\begin{align}
\frac{dv}{dt}  = a & =   a_{max}\left(1-\left(\frac{v}{v_d }\right)^\delta - \left(\frac{d^\star(v,\Delta v)}{d }\right)^2\right) \label{IDM1}\\
& d^\star(v,\Delta v)     =  d_0 + v{T_{set}} +\frac{v\Delta v }{2\sqrt{ba_{max}}} \label{IDM2}
\end{align}
Parameters of IDM to simulate different driver behavior are shown in Table \ref{tab:idmMobilParameters} \cite{volvo2}.

MOBIL algorithm is being used to decide when to change lanes in the simulator. It  makes a decision based on relative acceleration calculations regarding the following and lead vehicle in the current lane and the two adjacent lanes. In this context, with respect to the neighbouring vehicles, the following first safety criteria is  calculated:  

\begin{align}
\tilde{a}_n > b_{safe}   \label{safety_1}
\end{align} 
Here, $\tilde{a}_n$ refers to the new acceleration of the follower after making a lane change and $b_{safe}$ is the maximum safe deceleration. Safety criteria of the MOBIL guarantees accident free lane change decisions under the assumptions that other drivers react reasonably and there is no noise in the environment. If safety criteria is fulfilled, incentive criteria is calculated as following:
\begin{align}
\tilde{a}_e -a_e & + p (\tilde{a}_n - a_n) + q (\tilde{a}_o - a_o)  >  a_{th} \label{safety_2}
\end{align} 
where $\tilde{a}_e$, $\tilde{a}_n$ and $ \tilde{a}_o$ are the new accelerations, calculated by the IDM, for the lane changing, new follower and old follower vehicles, respectively. $a_e$, $a_n$ and $a_o$ refer to the current accelerations for the same vehicles. $p$ and $q$ are the politeness factor for the side and rear vehicles. $a_{th}$ is the lane change decision threshold. The parameters of MOBIL algorithm, that models different driver behaviors, are shown in Table \ref{tab:idmMobilParameters} \cite{volvo2}. 

MOBIL algorithm relies on a single threshold $a_{th}$ to make a decision if the decision is passed by the safety criteria. This is a main weakness of the algorithm since it is difficult to find an ideal threshold that may handle many different traffic situations and be robust to the measurement noise.  

\begin{table}[!t]
	\caption{IDM and MOBIL model parameters for different drivers.}
	\label{tab:idmMobilParameters}
	\centering
	\begin{tabular}{llrrr}
		& & Normal & Timid & Aggressive\\
		\hline
		Desired speed (m/s) & $v_\mathrm{set}$ & $25.0$ & $19.4$ & $30.6$ \\
		Desired time gap (s) & $T_\mathrm{set}$ & $1.5$ & $2.0$ & $1.0$\\
		Minimum gap distance (m) & $d_0$ & $2.0$ & $4.0$ & $0.0$\\
		Maximal acceleration (m/s\textsuperscript{2}) & $a_max$ & $1.4$ & $0.8$ & $2.0$\\
		Desired deceleration (m/s\textsuperscript{2}) & $b$ & $2.0$ & $1.0$ & $3.0$\\
		Politeness factor & $p$ & $0.05$ & $0.1$ & $0.0$\\
		Changing threshold (m/s\textsuperscript{2}) & $a_\mathrm{th}$ & $0.1$ & $0.2$ & $0.0$\\
		Safe braking (m/s\textsuperscript{2}) & $b_\mathrm{safe}$ & $2.0$ & $1.0$ & $3.0$\\
		\hline
	\end{tabular}
\end{table}

\subsection{Highway Simulation Details}

Simulation environment randomly generates scenarios out of initial conditions that are defined in Table \ref{table:Highway}.

\begin{table}[htb]
\caption{Highway Simulation Parameters} 
\vspace*{-2mm}
\centering    
\begin{tabular}{l c}   
\hline  \hline \vspace*{-2mm}&\\
Number of lanes, $n$   & $3 - 4$  \\
Number of vehicles, $m$   & $9 - 21$  \\
Maximum initial vehicle spread , $d_{long}$   & $200$ $m$ \\ 
Minimum inter-vehicle distance, $d_{\triangle}$  & $25$ $m$ \\ 
\vspace*{1mm}
Rear vehicles initial speed range, $[v^{rear}_{min}, v^{rear}_{max}]$ & $[15,25]$ $m/s$ \\
\vspace*{1mm}
Front vehicles initial speed range, $[v^{front}_{min}, v^{front}_{max}]$ & $[10,18]$ $m/s$ \\
\vspace*{1mm}
Initial speed range for ego vehicle, $[v^{ego}_{min}, v^{ego}_{max}]$ & $[10,15]$ $m/s$ \\
\vspace*{1mm}
Desired speed range for other vehicles, $[v^{d}_{min}, v^{d}_{max}]$ & $[18,26]$ $m/s$ \\
Desired speed for ego vehicle, $v^{d}_{ego}$ & $ 25 $ $m/s$ \\
Episode length, $d_{max}$ & $5000$ $m$ \\
\hline                                              
\end{tabular}
\label{table:Highway}                                
\end{table}

\subsection{Performance/Safety Indicators}

During the training and evaluation experiments, we monitor the number of accidents of the agents, average rewards of last the 100 episodes, number of lane changes, number of safety violations if the safety layer is integrated, ratio of successfully reaching terminal state which is in our case not the final destination but desired ego-velocity. Moreover, in order to monitor sample efficiency, we calculate the settling step referring to how many steps would take to reach \%95 of the settled reward. 

\section{RESULTS}

We have created two different benchmark scenarios to evaluate the influence of safety feedback, using Rainbow DQN, and compact state representation. Initial configuration is quite similar to our previous work \cite{itsc}: Ego-vehicle is surrounded by the 8 vehicle that shares normal driving behavior. We have trained Rainbow DQN, Rainbow DQN with safety and Double DQN agents over 1m training steps with three different seeds. In the second configuration, we took inspiration from \cite{volvo2} and increased the surrounding vehicle number to 20, uniformly sample different driver behaviors from \ref{tab:idmMobilParameters} and inject Gaussian measurement noise to the position and velocity of the vehicles. For this configuration, we trained three agents:
\begin{itemize} 
\item {Rainbow: Rainbow DQN without safety, with standard state representation (the information of all vehicles is provided)}
\item {Rainbow-blindpsot: Rainbow DQN with standard representation including blind-spot sensor and safety feedback} 
\item {Rainbow-blindspot-comp: Rainbow DQN with compact representation (only following and lead vehicles in the each lane provided) including blind-spot sensor and safety feedback}
\end{itemize}

\subsection{Noise Free Dynamic Highway Environment}

\begin{figure}[h]
    \centering
    \includegraphics[width=1\columnwidth]{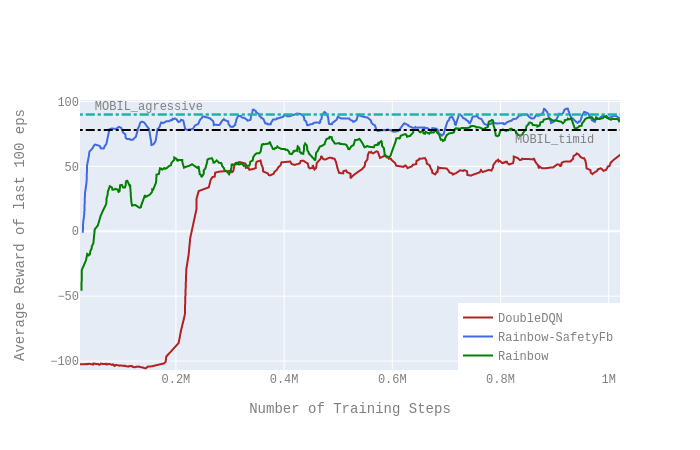}
    \caption{Training performance of three agents in seed a}
    \label{fig:easyenv}
\end{figure}

Figure \ref{fig:easyenv} shows the average reward of last the 100 episodes over the 1M training steps. As it can be clearly seen, rainbow DQN performs significantly better than double DQN in every aspect. We have also observed this superiority in other seeds and validation runs. Thus, we have not employed the double DQN to the more challenging scenarios. 

\begin{table}[h]
	\caption{Average reward over 1000 episodes} 
	\vspace*{-2mm}
	\centering    
	\begin{tabular}{l  c c c c}
	    & Solved &  mean reward & settling \\
	  Observations  & Eps. Ratio &  & step\\
	  \hline \hline \vspace*{-2mm}&\\
	  Rainbow & $96\%$ & $91$ & $1M$ \\
  	  Rainbow-blindspot & $99\%$ & $93 $ & $200k$\\
	  DoubleDQN & $70\%$ & $64$ &$1.5M$\\
	  MOBIL timid  & $83\%$ & $78$ & $0$\\
	  MOBIL aggressive  & $95\%$ & $90$ & $0$\\
	  \hline  
	\end{tabular}
	\label{table:results}
\end{table}

\subsection{Noisy, Uncertain Dynamic Highway Environment}

\begin{figure}[h]
    \centering
    \includegraphics[width=1\columnwidth]{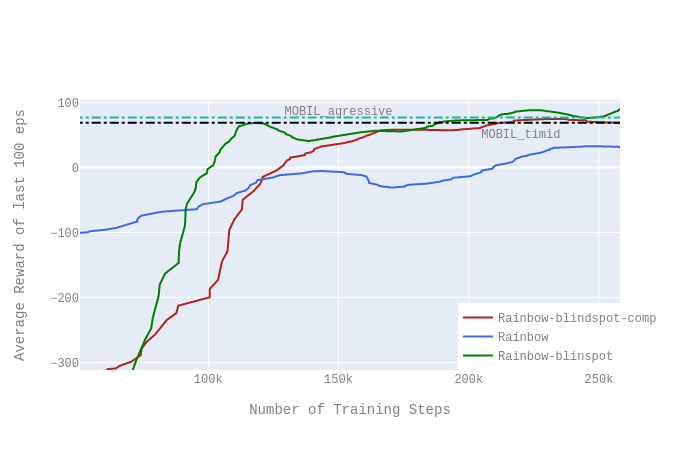}
    \caption{Training performance of three agents in seed a over 250k steps}
    \label{fig:safetyfb}
\end{figure}

Figure \ref{fig:safetyfb} shows average reward of last 100 episodes over the 250k training steps. As aligned with the previous findings, agents with safety feedback catch up performance of the MOBIL algorithm rapidly and surpass it after 200k time steps. Table \ref{table:results2} shows the performance in the evaluation run as well as training convergence. Moreover, performance of the agent with compact state representations is worse than the agent that uses states of the each vehicle whereas compact representation converges faster to the settling reward.

\begin{table}[h]
	\caption{Average reward over 1000 episodes} 
	\vspace*{-2mm}
	\centering    
	\begin{tabular}{l  c c c c}
	    & Solved &  mean reward & settling \\
	  Observations  & Eps. Ratio &   & step\\
	  \hline \hline \vspace*{-2mm}&\\
	  Rainbow & $82\%$ & $76 $ & $1M$\\
  	  Rainbow-blindspot & $92\% $ & $86 $ &$250k$\\
  	  Rainbow-blindspot-comp & $87\%$ & $82 $ &$180k$\\
	  MOBIL timid  & $72\%$ & $69$ & $0$\\
	  MOBIL aggressive  & $81\%$ & $77$ &$0$\\
\hline       
	\end{tabular}
	\label{table:results2}
\end{table}

\section{COMMENTS AND DISCUSSIONS}

\begin{figure}[h]
    \centering
    \includegraphics[width=1\columnwidth]{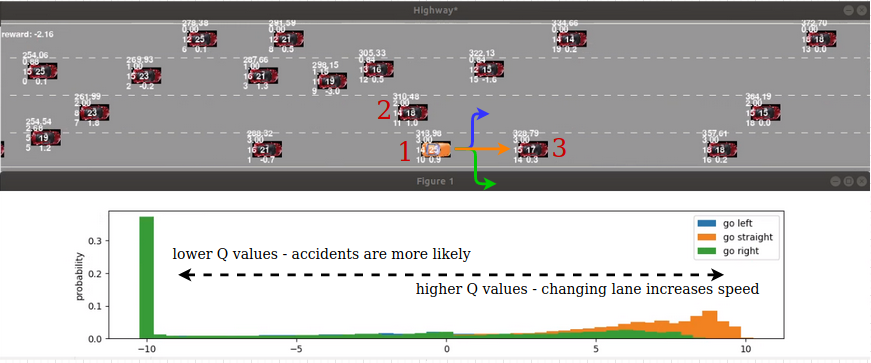}
    \caption{Probability distributions of Q values during the initial conditions. Agent is aware of going left causes an accident with the high probability. (Selected action: go straight)}
    \label{fig:trafic}
\end{figure}

In this section, firstly, we show different highway scenarios and try to understand the reasoning of the agent by using value distributions. In figure \ref{fig:trafic}, early moment of a challenging scenario is shown. The agent(orange car) is surrounded by the many vehicles and has only positive Q value expectations in the predicted Q value distributions, since going straight is always the safe state, while changing lanes is expected to cause either an accident or departure from the road. Consequently, agent waits until the adjacent left lane has enough space to overtake lead vehicle-3.

\begin{figure}[h]
    \centering
    \includegraphics[width=1\columnwidth]{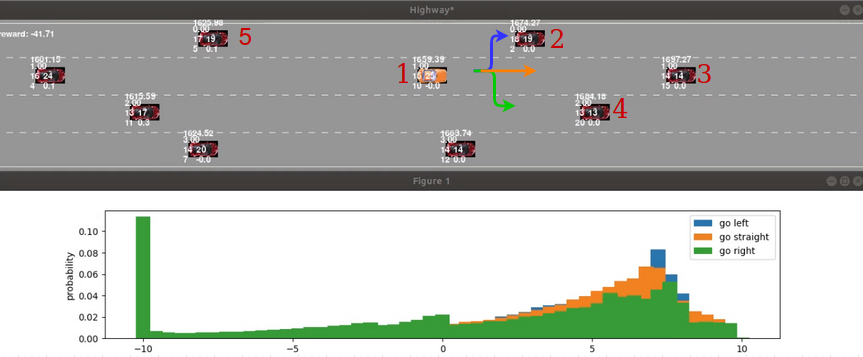}
    \caption{Probability distributions of Q values before changing lanes to the left. (Selected action: go left) }
    \label{fig:turnleft}
\end{figure}

In the second scenario shown in figure \ref{fig:turnleft}, probability distributions of value function is shown just before overtaking the slower vehicle-5 by changing lanes to the left. In this typical lane change scenario, agent accurately predicts the behavior of the following vehicle-5 in the new lane and executes lane change to the left.

\begin{figure}[h]
    \centering
    \includegraphics[width=1\columnwidth]{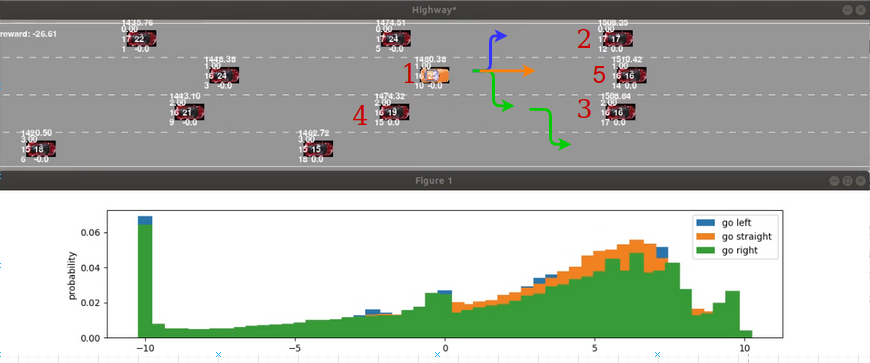}
    \caption{Probability distributions of Q values before changing lanes to the right. (Selected action: go right) }
    \label{fig:tactical}
\end{figure}

Figure \ref{fig:tactical} shows the interesting scenario where long term planning is required to make the right decision. Agent is blocked with the slower vehicle-5 and two adjacent lanes are also blocked with other agents (vehicle-2-3) where the adjacent left lead vehicle-2 is faster. Instead of selecting the greedy action and trying to turn left, agent selects to turn right and immediately changes its lane again to reach free lane in the bottom. 

Summary of the findings are as follows:

\begin{itemize}

\item{Rainbow DQN performed better than Double DQN in all handled test scenarios.}
\item{Measurement noise and randomized agents degraded performance of the both MOBIL and DRL agents}
\item{Integrating the proposed reward feedback from the safety layer to DRL resulted with top performance in the most challenging case of 20 vehicles. In the simpler scenarios, its deployment has accelerated the training, i.e. convergence speed.}
\item{Compact representation of the environment perform slightly worse than the full environment representation but training converges faster. }
\end{itemize}

\section{CONCLUSIONS AND FUTURE WORK}
In this paper, we have clearly demonstrated there is a promising potential of deploying tactical decision making algorithms for automated lane change functionality by using Rainbow DQN with safety layer that provides feedback to its reward function. We have shown through comparative studies that the deployment of Rainbow DQN integrated with a novel safety layer feedback significantly accelerated training dynamics in the developed realistic simulation environment. 

As for our future work, the proposed approach will be combined with learning from expert demonstrations approach to reach an even better sample efficiency.

\addtolength{\textheight}{-12cm}   





\end{document}